\pdfoutput=1

\documentclass[10pt,twocolumn,letterpaper]{article}

\usepackage{cvpr}
\usepackage{times}
\usepackage{epsfig}
\usepackage{graphicx}
\usepackage{amsmath}
\usepackage{amssymb}
\usepackage{gensymb}
\usepackage{multirow}

\usepackage[bookmarks=false]{hyperref}

\cvprfinalcopy 

\def\cvprPaperID{****} 

\begin{document}

\title{Exploring The Spatial Reasoning Ability of Neural Models in Human IQ Tests}

\author{
Hyunjae Kim\thanks{These authors contributed equally to this work.} \\
Department of Computer Science and Engineering, Korea University \\
{\tt\small hyunjae-kim@korea.ac.kr}
\and
Yookyung Koh\footnotemark[1] \\
NAVER\\
{\tt\small yookyung.koh@navercorp.com}
\and
Jinheon Baek\\
School of Computing, KAIST \\
{\tt\small jinheon.baek@outlook.kr}
\and
Jaewoo Kang\thanks{Corresponding author.} \\
Department of Computer Science and Engineering, Korea University\\
{\tt\small kangj@korea.ac.kr}
}

\maketitle

\begin{abstract}

Although neural models have performed impressively well on various tasks such as image recognition and question answering, their reasoning ability has been measured in only few studies.
In this work, we focus on spatial reasoning and explore the spatial understanding of neural models.
First, we describe the following two spatial reasoning IQ tests: rotation and shape composition.
Using well-defined rules, we constructed datasets that consist of various complexity levels.
We designed a variety of experiments in terms of generalization, and evaluated six different baseline models on the newly generated datasets.
We provide an analysis of the results and factors that affect the generalization abilities of models.
Also, we analyze how neural models solve spatial reasoning tests with visual aids.
Our findings would provide valuable insights into understanding a machine and the difference between a machine and human.

\end{abstract}

\begin{figure*}[t]
    \begin{center}
    \includegraphics[scale=0.55]{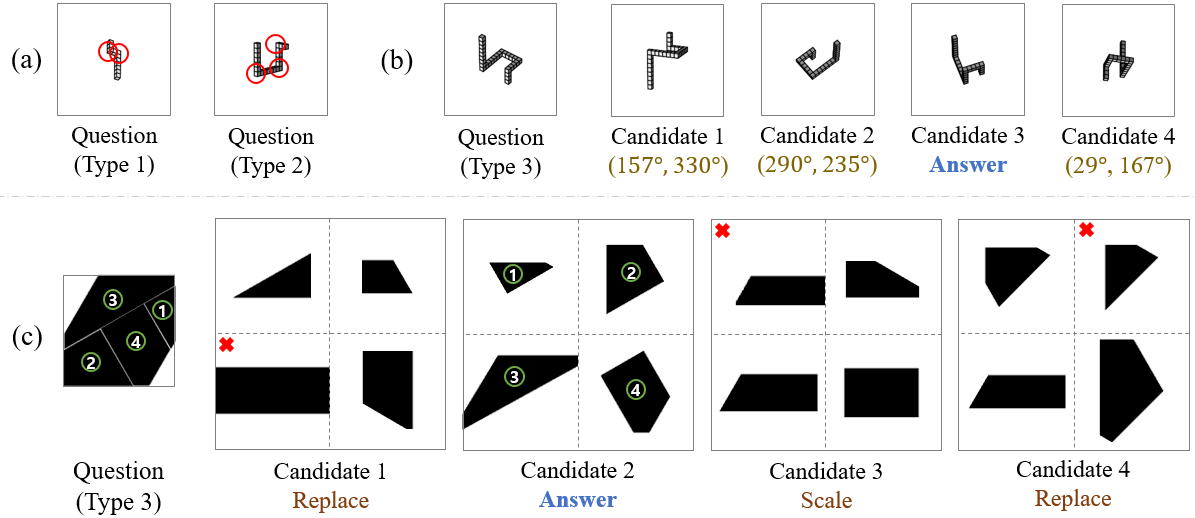}
    \caption{Data examples of the rotation task and shape composition task. 
    (a) shows that the shapes of images vary depending on complexity levels. 
    The left image in (a) shows a question image of a level 1 problem, and the right image shows a question image of a level 2 problem.
    Each edge of the objects in the images is joined by one shared joining block that is represented by the red-colored circles in (a).
    (b) depicts a level 3 problem in the rotation task.
    Comparing the question images in (a) with the question image in (b), the number of edges increases, which means the complexity of the problems increases.
    (c) illustrates an example of the shape composition task when the number of piece images is four ($m=4$).
    Note that there is no line in the original question images of the shape composition task.
    The numbers that are circled in green in the question image correspond to those in Candidate 2.
    The modified piece images are denoted by a red-colored "X" so that the original image is not produced.
    We scale a piece image (Scale) or replace a piece image with an another image (Replace).
    }
    \label{fig_data_example}
    \end{center}
\end{figure*}

\section{Introduction}
Over the last few years, deep learning approaches have been
a huge success in diverse domains.
Sophisticated systems even outperform humans in several tasks such as image recognition and question answering.
However, most tasks are focused on recognizing surficial patterns in data.
Also, only few works aim to solve tasks that require reasoning skills beyond pattern recognition or information retrieval.
Several recent works used human IQ tests to evaluate the reasoning abilities of machines \cite{wang2016solving}.
In the vision domain, several works utilized Raven’s Progressive Matrices (RPM) as a test bed for measuring the abstract reasoning abilities of neural models \cite{hoshen2017iq,barrett2018measuring,zhang2019raven}.
These tasks are challenging even for sophisticated systems because the systems are required to understand the logic of humans.

While following the spirit of existing works that dealt with a variety of reasoning, we focus on spatial reasoning.
We utilize human IQ tests, which are referred to as spatial reasoning tests, to explore the spatial understanding of neural models.
Spatial reasoning tests require to mentally visualize and transform objects in 2D or 3D spaces. 
For instance, to solve problems of rotation task in Figure 1(b), one should first mentally rotate objects in 3D space, and then determine whether the rotated objects are the same or not by visually comparing them. 
Hence, both image recognition and spatial comprehension are required to solve the task.

Of the various tasks in spatial reasoning IQ tests, we use the rotation task and the shape composition task.
The rotation task involves finding an object that is different from the 3D-polyomino in the given image, as shown in Figure \ref{fig_data_example}(b).
The rotation task involves recognizing visual features in 2D and visualizing rotated features in three dimensions.
The shape composition task, similar to solving a Tangram puzzle, involves choosing a set of pieces that would produce the given shape if combined, as shown in Figure \ref{fig_data_example}(c).
Hence, the shape composition task evaluates the ability to aggregate spatial information and understand the relative size, edges, and angles from given images.

A well-designed dataset is needed to evaluate the spatial reasoning ability of models.
However, spatial reasoning IQ tests are either unavailable due to the copyright issues or insufficient in the number of test samples.
To address this issue, we systematically generate a well-defined dataset.
The dataset is intentionally created simple to recognize its figures but challenging for reasoning.
We define three and four complexity levels for the rotation task and shape composition task, respectively.
In the rotation task, the complexity level is determined based on the shapes of objects in images whereas the complexity level in the shape composition task is determined based on the number of piece images in each candidate image. 
Creating different complexity levels enables our experiments to be expandable.
We generated 10,000 problems for complexity level. There is a total of 70,000 problems consisting of 750,000 images, which is sufficient for evaluating the reasoning ability of neural models.\footnote{We will make our data public after publication.}

Once a model learns a certain reasoning ability, the model should generalize the ability to unseen situations.
Thus, we evaluated models in the \textit{extrapolation} setting, as well as in the \textit{neutral} setting.
Unlike the neutral setting where the same complexity levels of problems appear in both the training and test sets, the extrapolation setting uses test sets that contains more complex problems.
Although it is easy for humans to apply their knowledge acquired from solving simple problems to more complex problems, it is challenging for machines to do so.

Finally, we describe a variety of baseline models that are commonly used in the visual reasoning domain.
For the shape composition task, we propose a novel model called CNN+GloRe, which is based on the GloRe unit introduced in \cite{chen2019graph}.
CNN+GloRe recognizes each piece image and then combines them in various ways like humans.
We provide the experimental results of all the  baselines.
From the results, we analyze factors that affect the generalization abilities of the models in spatial reasoning tests.
Also, we visualize which part of the input image has a strong signal for predicting the answer by a gradient-based approach. This explains the most important question: \textbf{how do neural models solve spatial reasoning tests?}
We believe our findings would be valuable insights into understanding a machine and the difference between a machine and human.


\section{Spatial Reasoning Tests}

Like verbal questions in \cite{wang2016solving} and RPMs in \cite{hoshen2017iq,barrett2018measuring,zhang2019raven}, spatial reasoning tests have been used and studied for decades in the fields of psychology, education, and career development \cite{gardner1992multiple,clements2004geometric,ehrlich2006importance}.
To the best of our knowledge, this work is the first to utilize spatial reasoning tests to study recent deep learning based models.
Various types of IQ tests differ in their goal.
The verbal questions measure an understanding of the meaning of words.
RPMs measure the ability to find abstract rules or patterns such as progression, AND, OR and XOR, from the given context images.
On the other hand, spatial reasoning tests measure the ability to mentally visualize and transform objects in 2D or 3D spaces.

Spatial reasoning is not new in the vision domain because various tasks such as movement prediction and the room-to-room navigation task implicitly require spatial reasoning \cite{suchan2019out,anderson2018vision}.
However, they focused on their main tasks and did not explicitly study on spatial reasoning.
On the other hand, our main tasks are to solve spatial reasoning tests, which implies we focus on spatial reasoning itself.
Similar to our work, the CLEVR dataset was proposed to study visual reasoning \cite{johnson2017clevr}. 
However, spatial relationships that CLEVR requires are directly captured in given 2D images, without obtaining hidden information in the given images. 
Our aim is to determine whether models can obtain hidden spatial information from superficial visual inputs.

\subsection{Rotation}
\subsubsection{Task Description}
In the rotation task, a 3D-polyomino is an object in an image, which is formed by joining one block edge to edge.
A model has to choose the correct answer out of four candidate answers which has a different 3D-polyomino object from the given question.
The remaining three candidate answers are made by rotating the polyomino in the question image both horizontally and vertically in 3D space.
All edges are 90 degrees to each other.
Figure \ref{fig_data_example}(b) shows an example of the rotation task.

\subsubsection{Related Work}
In the field of vision systems, learning rotation was used as tools for other downstream tasks such as image classification \cite{gidaris2018unsupervised}.
There exists three main approaches for learning rotation-invariant representations in downstream tasks.
The first approach involves predicting the degree of rotation \cite{kanezaki2018rotationnet,chen2019self}, but it requires target labels for the degrees of rotation.
The second approach is a structural method which involves transforming kernels in CNN layers to obtain rotation invariant features from images \cite{shen2017patch,weiler2018learning}. 
However, this approach considers only the 2D rotation of 2D images.
In our task, there are no labels for the degrees of rotation, and the 3D rotation of objects from 2D images is considered.
Hence, the last approach, using the vector distance, is the most suitable for our task. 
Like \cite{koch2015siamese,cheng2016rifd}, we train our model on the vector distance between question and candidate answers.

\subsection{Shape Composition}
\subsubsection{Task Description}
The shape composition task involves choosing the correct set of piece images that would produce the original image if combined, as shown in Figure \ref{fig_data_example}(c).
Unlike the rotation task, each candidate answer in the shape composition task contains $m$ number of pieces, which ranges from 2 to 5. 
The correct answer is the candidate with these $m$ pieces that can produce the original image when combined. 
The remaining three incorrect candidate answers have pieces that are not part of the original image.

\subsubsection{Related Work}
Compared to shape composition tasks in the previous works \cite{pich2018logical}, in our shape composition task, only images are given without any information such as the lengths of sides of objects.
The shape composition task is similar to solving a jigsaw puzzle in that all pieces are combined to produce the original image.
Methods and skills used for solving jigsaw puzzles can be applied to biology \cite{marande2007mitochondrial}, archaeology \cite{paumard2018image}, image editing \cite{sholomon2014generalized}, and learning visual representations \cite{doersch2015unsupervised,noroozi2016unsupervised,santa2017deeppermnet,carlucci2019domain}.
Most existing works divide an original image into a grid of equal-sized squares, and then assemble the square pieces to produce the original image.
In \cite{gur2017square}, the shape of pieces was converted to a rectangle.
On the other hand, we assemble polygons with various shapes.

\begin{table}[]
\centering
\footnotesize
\begin{tabular}{|c|c|c|}
\hline
Tasks   & Complexity Levels & Statistics  \\ \hline
Rotation     & \begin{tabular}[c]{@{}c@{}}Three levels\\ depending on\\ the number of \\ edges k\end{tabular}       & \begin{tabular}[c]{@{}c@{}}7k/1k/2k problems\\ for each level,\\ 30k problems in total\end{tabular} \\ \hline
Shape Composition & \begin{tabular}[c]{@{}c@{}}Four levels\\ depending on\\ the number of \\ piece images m\end{tabular} & \begin{tabular}[c]{@{}c@{}}8k/1k/1k problems\\ for each level,\\ 40k problems in total\end{tabular} \\ \hline
\end{tabular}
\caption{Summary of the complexity levels and statistics of our datasets.
}
\label{tab:settings}
\end{table}

\begin{figure*}[h]
    \begin{center}
    \includegraphics[scale=0.55]{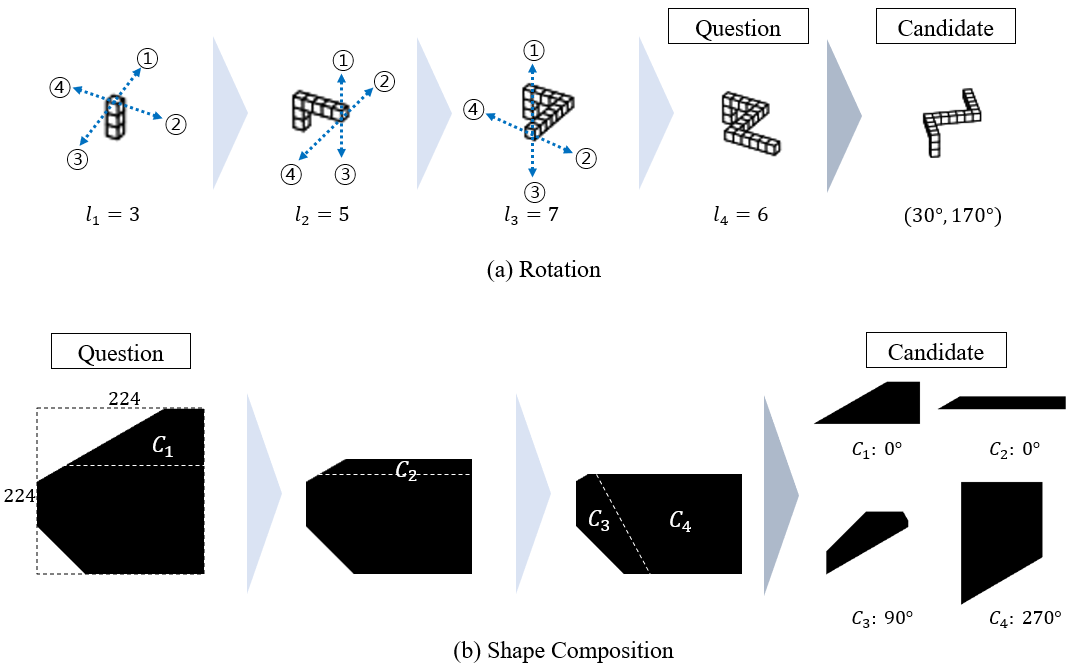}
    \caption{Description of data generation steps.
    (a) This is a case where the object has 4 edges ($k=4$). A set of edge lengths is $l=(3,5,7,6)$. Blue dotted arrows denote the perpendicular directions to which the next edge can go. At each step, the direction is randomly selected and the edge is joining the last block. A candidate answer is generated by rotating the question image.
    (b) This is a case where the number of pieces is 4 ($m=4$). The area of the original image is larger than 25000. It is linearly cut at each step. Each piece $C_i$ is larger than 3000.
    Each piece is randomly rotated at 0\degree, 90\degree, 180\degree, or 27 \degree clockwise.
    }
    \label{fig:datagen}
    \end{center}
\end{figure*}

\section{Data Construction and Experimental Settings}
In this section, we describe the data generation process for each task.
We focus on measuring reasoning ability, and not recognition ability.
Hence, we use basic blocks and polygons and create incorrect candidate answers not too different from the question and the correct answer.
Each image in our datasets has 224x224 pixels so that the images can be easily viewed by humans.
The image size can be reduced for memory efficiency.

A problem in both tasks consists of one question and four candidate answers including the correct answer.
We classify problems into three complexity levels for the rotation task and four complexity levels for the shape composition task.
Description of complexity level is in the subsections below.
Utilizing the complexity levels, we conduct various experiments in terms of generalization.
All the experiments are categorized into the neutral and extrapolation settings.
In the neutral settings, the training and test sets contain the same complexity levels of problems.
On the other hand, in the extrapolation setting, we use test sets with more complex problems.
Table \ref{tab:settings} describes complexity levels and data statistics for each task.

We denote our experiments as $|TR \rightarrow TE|$ where $TR$ denotes complexity levels that the training set contains and $TE$ denotes complexity levels that the test set contains.
For example, $|1,2 \rightarrow 3|$ denotes that level 1 and level 2 problems appear in the training set and level 3 problems appear in the test set. In the neutral setting, $|all\rightarrow all|$ denotes all complexity levels are used in both training and test sets.
Problems of each complexity level have the same probability of being selected.

We evaluate the baseline models on the following two test sets: 1) An in-distribution (in-dist) test set with the same complexity levels as the training set, 2) An out-of-distribution (out-dist) test set with different complexity levels from the training set.
The validation set is used only for tuning the hyper-parameters of each model.\footnote{Previous studies reported the performance of models on only validation and test sets \cite{barrett2018measuring,zhang2019raven}.
However, it is not fair to compare performance on the validation set with that on the test set since models can fit to the validation set.}
We use accuracy as the evaluation metric.

\subsection{Rotation}
In this section, we describe how to systematically create a dataset for the rotation task.
First, we denote sets of edge lengths, angles, and directions as $L, A$ and $D$, respectively.
The edge length $L$ is equal to the number of blocks that form each edge, and each edge can have 3 to 9 blocks.
$A$ is the set of rotation angles, including vertical and horizontal angle pairs.
The angles can be any degrees in the intervals of $[\theta + 15\degree, \theta + 75\degree)$ where $\theta$ is a multiple of $90\degree$.
We exclude angles that are $15\degree$ apart from the angles that are a multiple of $90\degree$ since images are indistinguishable using the angles in this interval.
We create an object by joining each edge, depending on the four directions in which the last edge can be 90 degrees to the previous edge.
$D$ is a set of directions where each element represents one of four possible perpendicular directions of the last edge.
Finally, a polyomino is expressed as a triplet $(l, \alpha, \delta)$ where $l = [l_1,...,l_k], \alpha = [\alpha_v,\alpha_h]$ and $\delta = [\delta_1,...,\delta_{k-1}]$.
Elements of $l$ and $\delta$ are sampled from $L$ and $D$, respectively.
$\alpha_v$ and $\alpha_h$ sampled from $A$ are vertical and horizontal angles, respectively.
$k$ denotes the number of edges of an object in images.

A question and three candidate answers have $l$ and $\delta$ in common.
The question and the correct answer have the same $l$ and have different $\delta$s, which results in different polyominoes.
$\alpha$s of the question image and four candidate answer images are taken from four different intervals, which implies that two out of the five images share the same interval either for $\alpha_v$ or $\alpha_h$.

There are three complexity levels in the rotation task.
The complexity levels are determined depending on the number of edges, $k\in [3,4,5]$.
A larger $k$ produces more complex problems.
For each level, we randomly generate 10,000 problems from combinations of $(l, \alpha, \delta)$, which results in a total of 30,000 problems.
7K, 1K, and 2K problems are used for the training, validation, and test sets, respectively.

In the case of humans, it does not take much effort to acquire reasoning skills.
Hence, our datasets are smaller than the benchmark datasets used in the computer vision field, but the size of our datasets is sufficient for studying reasoning.
An excessive amount of data can cause unintended bias or contain irrelevant features, which makes it difficult to train models.

\subsubsection{Experimental Settings}
We conduct one experiment in the neutral setting ($|all\rightarrow all|$), and three experiments in the extrapolation setting for the rotation task, which are denoted as follows: $|1\rightarrow 3|$, $|2\rightarrow 3|$ and $|1,2\rightarrow 3|$.

We use 7K, 1K, 1K problems for the training, validation and in-dist test sets, respectively, in the neutral setting and use additional 1K problems for out-dist test sets in the extrapolation setting.
The problems are sampled from the corresponding problem sets of each complexity level.
In the extrapolation setting, the same out-dist test set is used.

\subsection{Shape Composition}

For dataset for the shape composition task, we first created original images.
We generated $224\times 224$ initial images coloured in black.
We linearly cut the initial images twice.
We use an initial image as an original image only if its size is larger than 25,000, where the size denotes the number of remaining pixels coloured in black after cutting linearly.
Here, the linear operation has the following two requirements: 1) The slope of the cutting line is either $0\degree, 30\degree, 45\degree$ or $60\degree$, 2) The point corresponding to the y-intercept is between 1/4 and 3/4 of the image height.

\begin{figure*}[h]
    \begin{center}
    \includegraphics[scale=0.65]{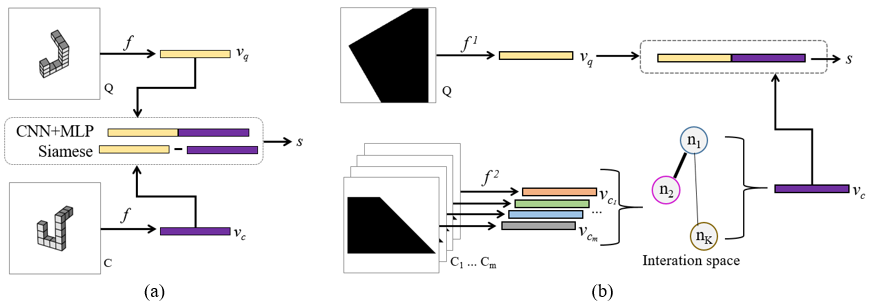}
    \caption{Illustration of the following models used for spatial reasoning tasks. (a): CNN+MLP and Siamese for the rotation task. (b): CNN+GloRe for the shape composition task. $f^1$ is a recognizer for question images and $f^2$ is a recognizer for piece images.
    }
    \label{fig:models}
    \end{center}
\end{figure*}

We then linearly cut the original image into $m$ number of pieces. 
When $m$ is 2, 3, or 4, the size of each piece is in the range of 3,000 to 30,000 and when $m=5$, the size of each piece is in the range of 2,000 to 30,000.
The cut pieces are then rotated by $0\degree, 90\degree, 180\degree$ or $270\degree$, but never flipped.
This set of $m$ number of pieces cut from the original image forms the correct answer.
For two of the incorrect candidate answers, $m$ pieces are also from the same original image but only one piece is randomly replaced with another piece of similar size.
For the remaining incorrect candidate answer, one of the pieces of the correct answer is scaled, which makes it impossible to form the original image.

In the shape composition task, there are four complexity levels.
The complexity levels are determined depending on the number of piece images $m\in [2,3,4,5]$.
A larger $m$ produces more complex problems.
For each level, we randomly generate 11,000 problems, which results in a total of 44,000 problems.
8K, 2K and 1K problems are used for the training, validation and test sets, respectively.

\subsubsection{Experimental Settings}
In the shape composition task, we conduct one experiment in the neutral setting as in the rotation task.
Also, we conduct two groups of seven experiments in the extrapolation setting, where the same out-dist test sets are used in the same group.
In the first group, the complexity levels of the test set is 4: $|2\rightarrow 4|$, $|3\rightarrow 4|$, $|2,3,\rightarrow 4|$ and $|1,2,3\rightarrow 4|$. In the second group, the complexity levels of the test set are 3 and 4: $|1\rightarrow 3,4|$, $|2\rightarrow 3,4|$ and $|1,2, \rightarrow 3,4|$.
We use 7K, 1K, 1K and 1k problems for the training, validation, in-dist test and out-dist test sets, respectively.

\section{Model Architectures and Training}

In this section, we describe the baseline models used in spatial reasoning tasks.
All of the baselines are neural networks that are commonly used in the field of vision reasoning or designed to be task-specific.

In both the rotation and shape composition tasks, an image of one question and images of four candidates are given as inputs.
Models have to calculate a similarity score $s$ for each question and candidate answer pair $(\mathrm{Q}, \mathrm{C})$.
A Softmax function is applied to the scores of four pairs representing the probabilities of each candidate answer being the correct answer.
For optimization, cross-entropy loss is used.

\subsection{Rotation}
Given the image pair $\mathrm{Q}$ and $\mathrm{C}$, we first encode this image pair into vector representations using an image recognizer $f$ as follows: $v_q = f(\mathrm{Q})$ and $v_c = f(\mathrm{C})$.
Next, we combine $v_q$ and $v_c$, and compute the score of the pair as follows: $s = g(f(Q), f(C))$ where $g$ is a reasoning module that captures relationships between questions and candidate answer images, and then reason the correct answers.
The structure of the baseline models for the rotation task are determined depending on $f$ and $g$.

\subsubsection{CNN+MLP} Similar to \cite{hoshen2017iq}, we used a 4-layer CNN as the image recognizer.
First, the CNN encodes input images into vectors.
The vectors are concatenated to each other, and then fed into a 2-layer MLP to compute a similarity score $s$.
Figure \ref{fig:models}(a) illustrates the CNN+MLP.

\subsubsection{ResNet+MLP} 
We compare the 4-layer CNN with the deeper CNN ResNet-50 \cite{he2016deep} which is one of the most widely used image recognizers.
Like CNN+MLP, ResNet+MLP has two MLP layers.

\subsubsection{Siamese} 
Using the vector distance between two images is one way to solve the rotation task.
This approach can be used when the degree of rotation is not provided.
Using Siamese networks \cite{koch2015siamese}, the vector distance between $v_q$ and $v_c$ is calculated by cosine similarity.\footnote{We trained Siamese using the L1 loss, but the performance decreased.}
Unlike CNN+MLP and ResNet+MLP, our Siamese model is optimized by the binary cross entropy losses and the four similarity scores of each question and candidate pair. 
When evaluating Siamese, we choose the candidate answer with the lowest similarity score as the correct answer.
Figure \ref{fig:models}(a) illustrates the Siamese model.

\subsection{Shape Composition}
Unlike the rotation task, in the shape composition task, a candidate image consists of multiple piece images $\mathrm{C}_1,...,\mathrm{C}_m$ where $m \in [2,3,4,5]$.
Thus, we have to consider a model to aggregate piece vectors $v_{c_1},...,v_{c_m}$ and represent a candidate vector $v_c$ with a fixed length.
We compare models with different aggregation functions.

\subsubsection{CNN+MLP} We concatenate piece vectors $v_{c_1},...,v_{c_m}$ encoded using CNN, and feed the concatenated vectors to the MLP layers to obtain a fixed sized candidate vector $v_c$.
The number of hidden units in the MLP layer is set to the maximum number of pieces, and we randomly feed each image to each unit.

\subsubsection{CNN+Max}
Max-pooling is a simple function that calculates dimension-wise maximum values of piece vectors.
The MLP layers are used to compute a score.

\subsubsection{CNN+GloRe} Following \cite{chen2019graph}, we adopt Global Reasoning unit (GloRe unit) to create a novel model, CNN+GloRe, that is specified to the shape composition task.
The CNN+GloRe is illustrated in Figure \ref{fig:models}(b).
The GloRe unit was proposed to capture not only local but also global relationships between image regions.
The GloRe unit maps image regions in a coordinate space into nodes in an interaction space, then operates weighted graph pooling \cite{kipf2016semi}.
We combine the GloRe unit with CNN for our shape composition task, where each piece image is regarded as an image region.
The GloRe aggregates piece vectors into a candidate vector $v_c$ as follows:
\begin{equation}
v_c = h(f(C_{1}),...,f(C_{m}))
\end{equation}
where $h$ is the GloRe unit.
$v_q$ and $v_c$ are then concatenated and fed into the MLP layers.\footnote{We implemented GloRe using the code released by the authors\footnote{https://github.com/facebookresearch/GloRe}.}

In detail, each node in an interaction space is obtained as follows:
\begin{equation}
n_i = \sum^{m}_{j=1} b_{ij}v_{c_j}
\label{eq_compute_node}
\end{equation}
where $b_i \in \mathcal{R}^{m}$ is a learnable parameter and $j$ is the index of a piece image of the candidate answer.
Unlike in the rotation task, the recognizer for encoding questions and the recognizer for piece images are different. 
The question and piece vectors are encoded as $v_{q} = f^{1}(Q)$ and $v_{c_j} = f^{2}(C_j)$, respectively.
All the nodes $N=[n_1,...,n_K]^T$ in the graph convolutional layers are as follows:

\begin{equation}
Z = (I-\mathcal{A}) \cdot tanh(\mathcal{A} N G^1)G^2
\end{equation}
where $I \in \mathcal{R}^{K \times K}$ is an identity matrix and $\mathcal{A} \in \mathcal{R}^{K \times K}$ is a trainable adjacency matrix that is randomly initialized.
$G^1 \in \mathcal{R}^{d \times d}$ and $G^2 \in \mathcal{R}^{d \times d}$ are trainable weights where $d$ is the dimension size of piece vectors.
$Z \in \mathcal{R}^{K \times d}$ is the representation of the graph convolution output nodes.

Similar to Equation \ref{eq_compute_node}, the obtained node representations in the interaction space are mapped to the coordinate space as follows:

\begin{equation}
\ddot{v}_{c_i} = \sum^{K}_{j=1} e_{ij}z_{j}
\end{equation}
where $z_{j}$ is the $j$-th row of $Z$, and $e_i \in \mathcal{R}^{m}$ is a learnable parameter.
Finally, a candidate vector $v_c$ is computed as the element-wise mean of vectors $\ddot{v}_{c_1},...,\ddot{v}_{c_m}$.
Figure \ref{fig:models}(b) illustrates the graph-based structure of the CNN+GloRe model.

\subsection{Hyperparameter Settings}
In this section, we summarize all the hyper-parameter settings of the baseline models.
We implemented baseline models using PyTorch.\footnote{https://pytorch.org}
The source code for reproduction is publicly available at \textit{github.com/blind}.

\subsubsection{Rotation}
For CNN, we used 4 convolutional layers each of which has $16,32,64,128$ feature maps, respectively.
The kernel size is set to 7 for all layers. 
The dimension $d$ of the image vectors is set to 512. 
We used a 2-layer MLP with a ReLu activation function for CNN+MLP.
The same image recognizer is used for question and candidate images in the rotation task.
CNN+MLP and Siamese were optimized by an SGD optimizer with an initial learning rate of 0.1. 
The batch size is set to 64. 
ResNet+MLP was optimized by the Adam optimizer \cite{kingma2014adam} with an initial learning rate of 0.0005, and the batch size was set to 16.
A learning rate decay of 0.9 at each epoch was used.

\subsubsection{Shape Composition}
In the shape composition task, we used the same hyperparameters for CNN, which were used in the rotation task.
However, we used two CNNs: one for question images and the other for candidate images.
For CNN+GloRe, we used the same GloRe unit used in \cite{chen2019graph}.
The SGD optimizer is used for CNN+GloRe, and the Adam optimizer is used for the other models.
A learning rate of 0.1 is used for the SGD optimizer and a learning rate of 0.0005 is used for the Adam optimizer. 
A batch size of 64 is used for all models.

\begin{table*}[]
\centering
\footnotesize
\begin{tabular}{cc|c|c|c|c|c|c|c|c|c|c|c|c|}
\cline{3-14}
 &  & \multicolumn{4}{c|}{CNN+MLP} & \multicolumn{4}{c|}{ResNet+MLP} & \multicolumn{4}{c|}{Siamese}  \\ \hline
\multicolumn{2}{|c|}{Settings}  & Val    & In    & Out    & Gap    & Val    & In    & Out    & Gap    & Val     & In    & Out    & Gap     \\ \hline
\multicolumn{1}{|c|}{Neutral}   &  $all \rightarrow all$   & 77.7    & \textbf{77.9}   & - & - & 52.5 & 52.4  & -   & - & 73.2 & 73.4 &- & -  \\ \hline
\multicolumn{1}{|c|}{\multirow{3}{*}{Extra}} & $1 \rightarrow 3$ &   96.2	  &  96.7       &  	32.8      &    63.9    &   53.2	   &  50.0    &   	\textbf{51.8}       &    -1.8    &    90.5    &  	91.6	  & 38.7   & 52.9 \\ \cline{2-14} 
\multicolumn{1}{|c|}{} & $2 \rightarrow 3$ &   82.5   &  79.1      & 	\textbf{72.7}    & 6.4  & 55.2     & 53.6	  &  52.3         &  1.3 &   78.0     & 	74.9  &    	64.1      &  10.8   \\ \cline{2-14} 
\multicolumn{1}{|c|}{}  & $1,2 \rightarrow 3$ &  87.3 & 86.0 & \textbf{71.8}  &  14.2  & 54.2 & 51.4 & 50.9     & 0.5 & 86.1 &  81.8 & 62.9 & 19.0 \\ \hline
\end{tabular}
\caption{Experimental results in the rotation task. Val, In and Out denotes performance on validation, in-dist and out-dist sets, respectively. Gap denotes the difference in performance between in-dist and out-dist.}
\label{tab:rotation_result}
\end{table*}

\begin{table*}[]
\centering
\footnotesize
\begin{tabular}{cc|c|c|c|c|c|c|c|c|c|c|c|c|}
\cline{3-14}
 &   & \multicolumn{4}{c|}{CNN+MLP} & \multicolumn{4}{c|}{CNN+Max} & \multicolumn{4}{c|}{CNN+GloRe} \\ \hline
\multicolumn{2}{|c|}{Settings}  & Val    & In   & Out   & Gap   & Val    & In   & Out   & Gap   & Val    & In    & Out    & Gap   \\ \hline
\multicolumn{1}{|c|}{Neutral} & $all \rightarrow all$ & 78.1 & 74.7 & - & - & 74.6 & 72.6 & - & -    & 79.7 & \textbf{77.2} & - & -    \\ \hline
\multicolumn{1}{|c|}{\multirow{5}{*}{Extra}} & $2 \rightarrow 4$ &   81.9 & 79.5 & 35.3 & 44.2 & 78.2 & 77.1 & \textbf{59.3} & 17.8  & 83.4  & 81.8  & 31.9  & 49.9  \\ \cline{2-14} 
\multicolumn{1}{|c|}{} & $3 \rightarrow 4$ &   79.1 & 76.1 & 52.4 & 13.7 & 72.8 & 71.6 & \textbf{66.5} & 5.1 & 80.8 & 81.2 & 50.7 & 28.5 \\ \cline{2-14} 
\multicolumn{1}{|c|}{} & $2,3 \rightarrow 4$  & 79.6 & 77.6 & 70.5 & 7.1 & 75.4 & 75.7 & 65.2 & 10.5 & 79 & 78.5 & \textbf{72.1} & 6.4 \\ \cline{2-14} 
\multicolumn{1}{|c|}{} & $1,2,3 \rightarrow 4$  & 79.7 &  79.2  & \textbf{70.1} & 9.1 & 79.6 & 77.9 & 64.9 & 13  & 82.2 & 80.6 & 69.8 & 10.8    \\  \cline{2-14} 
\multicolumn{1}{|c|}{} & $ 1 \rightarrow 3,4$ & 91.9 & 90.7 & 33.2 & 57.5  & 92.4 & 91.7 & \textbf{36.9} & 54.8  & 92.4 & 92.6 & 30.9 & 61.7      \\ \cline{2-14} 
\multicolumn{1}{|c|}{} & $ 2 \rightarrow 3,4$ & 82 & 80.5 & 51.4 & 29.1  & 80.9 & 79.6 & \textbf{64.9} & 14.7  & 83.9 & 81.5 & 48.1 & 33.4      \\ \cline{2-14} 
\multicolumn{1}{|c|}{} & $1,2 \rightarrow 3,4$  & 83.9 & 82.1 & \textbf{68.6} & 13.5 & 82.9 & 84 & 61.8 & 22.2 & 84.6 & 82.5 & 64.2 & 18.3    \\ \hline 
\end{tabular}
\caption{Experimental results on the organization task. Val, In and Out denotes performance on validation, in-dist and out-dist sets, respectively. Gap denotes the difference in performance between in-dist and out-dist.}
\label{tab:org_resutls}
\end{table*}

\section{Experimental Results}
This section discusses the experimental results and provides analysis.
Table \ref{tab:rotation_result} and Table \ref{tab:org_resutls} show that the results in the neutral and extrapolation settings for each task, respectively.
The overall section provides analysis of the experimental results which are consistent in both tasks.
In the rotation and shape composition section, we analyze the experimental results of the baseline models and focus mainly on the differences in their architecture.

\subsection{Overall}

Training on complex problems is more effective than training on simple problems.
Table \ref{tab:rotation_result} shows that performance of CNN+MLP in $|2 \rightarrow 3|$ is 39.9\% higher than that in $|1 \rightarrow 3|$.
Also, Table \ref{tab:org_resutls} shows that its performance in $|3 \rightarrow 4|$ is 48.0\% higher than that in $|2 \rightarrow 4|$, and performance in $|2 \rightarrow 3,4|$ is 54.8\% higher than that in $|1 \rightarrow 3,4|$.

Training on different complexity levels improves generalization.
Before we conducted experiments in the extrapolation setting, we predicted that the performance on out-dist test sets would increase if models were trained on various complexity levels.
In the shape composition task, our prediction is consistent with the following results: performance of CNN+MLP and CNN+GloRe in $|2,3 \rightarrow 4|$ is higher than that in $|3 \rightarrow 4|$, and performance in $|1,2 \rightarrow 3,4|$ is higher than that in $|2 \rightarrow 3,4|$.
However, the results from Table \ref{tab:rotation_result} ($|2 \rightarrow 3|$ and $|1,2 \rightarrow 3|$) and Table \ref{tab:org_resutls} ($|2,3 \rightarrow 4|$ and $|1,2,3 \rightarrow 4|$) are inconsistent with our prediction.

\begin{figure}[t]
    \begin{center}
    \includegraphics[scale=0.45]{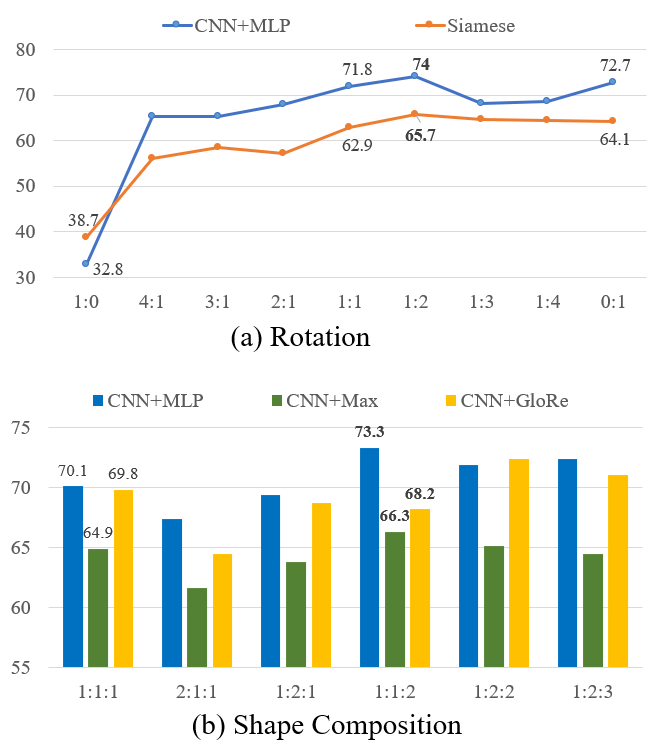}
    \caption{Results of the extrapolation tests with different ratios of complexity levels. (a): The results of the rotation test $|1,2\rightarrow 3|$.
    The ratios of complexity level 1 and level 2 problems are plotted on the x-axis.
    (b): The results of the shape composition test $|1,2,3\rightarrow 4|$.
     The ratios of complexity level 1, level 2, and level 3 problems are plotted on the x-axis.
    The performance on the out-dist test set is plotted on the y-axis.}
    \label{fig:result_ratio}
    \end{center}
\end{figure}

We hypothesized that adding an equal number of simple and complex problems to a training set would result in models learning fewer complex problems.
we conduct an additional experiment using different ratios of complexity levels in the training set.
We provide the result of $|1,2\rightarrow 3|$ for the rotation task, and $|1,2,3\rightarrow 4|$ for the shape composition task.
As Figure \ref{fig:result_ratio} shows, changing the ratio of complexity levels in the training set affects performance on the test set.
Figure \ref{fig:result_ratio}(a) shows that CNN+MLP and Siamese achieved the highest and second highest accuracy performance of 74.0\% and 65.7\%, respectively, when the ratio is 1:2 in the rotation task.
The performance of CNN+MLP and Siamese was 3.06\% and 4.45\%, respectively, higher than when training with the ratio of 1:1.
However, when simple problems are added (the ratios of 1:3 and 1:4), performance slightly decreases.
When the proportion of simple problems in the training set is too small, the simple problems may act as noise.
When the ratio of the simple problems increases, performance improves.
After the performance peaks with the ratio of 1:2, it decreases as the proportion of simple problems increases.
If the proportion of simple problems is too high, performance may not improve.
Figure \ref{fig:result_ratio}(b) shows the result of the shape composition test $|1,2,3\rightarrow 4|$.
The CNN+MLP model achieves the highest performance of 73.3\% when the ratio is 1:1:2, which shows that performance in both tasks improves as the proportion of more complex problems increases.
Thus, we confirmed that training on different complexity levels improves generalization.

\subsection{Rotation}

Siamese does not generalize well in most experiments.
In previous studies, the vector distance between two images was commonly used to learn rotation-invariant representations when the degree of rotation is not available.
Siamese networks, trained on the vector distance, generalized well on new situations \cite{koch2015siamese}.
However, in our experiments, Siamese performed relatively poor than CNN+MLP in most experiments as Table \ref{tab:rotation_result} shows.
These results implies that using only a recognizer is not enough to solve the rotation task.
A reasoning module such as MLPs that aggregates question and candidate answers is helpful.

A larger model size does not guarantee higher performance.
As shown in Table \ref{tab:rotation_result}, validation and test performances of ResNet+MLP are relatively low when training the ResNet+MLP from the scratch on our rotation task.
Since the low performance may be due to underfitting, we replaced ResNet with a pretrained ResNet and trained all weights in the model (ResNet-pre).
Though the pretrained ResNet is trained on the large amount of data in ImageNet dataset, ResNet-pre achieved the performances of 54.4\%, 52.0\% in validation and in-dist test set, respectively, in the neutral setting.\footnote{ResNet+MLP also performed relatively poor in the shape composition task although it requires more memory and training time.}
From these results, we conclude that a large model size does not always improve performance on spatial reasoning tasks.
Using heavy image recognizers can cause overfitting.
Instead, proper structures or training methods should be discussed to solve the spatial reasoning tasks.

\subsection{Shape Composition}

CNN+GloRe always outperforms other baselines in the neutral setting, but not in the extrapolation setting.
In the neutral setting, CNN+GloRe obtained 2.5\% and 4.6\% higher performance than CNN+MLP and CNN+Max, respectively, as shown in Table \ref{tab:org_resutls}.
However, in the extrapolation setting, CNN+MLP generally outperforms CNN+GloRe.
We assumed that CNN+GloRe would be able to learn shape composition skills (e.g., combining piece images) like humans, and achieve high performance in all experimental settings, but CNN+GloRe did not.
Since CNN+GloRe relies on features from images in training data, it obtained low performance in the extrapolation setting.
We believe that more work should be studied for developing models that perform in a similar way that humans solve Tangram puzzles, such as CNN+GloRe.
Also, this work should involve careful consideration of learning the principle of reasoning.

CNN+Max generalize well when trained on only one complexity level.
Table \ref{tab:org_resutls} shows that CNN+Max outperforms the other baselines in the following four experiments in the extrapolation setting: $|2 \rightarrow 4|$, $|3 \rightarrow 4|$, $|1 \rightarrow 3,4|$ and $|2 \rightarrow 3,4|$.
CNN+Max learns to find the most noticeable features from images, regardless of the number of piece images, resulting in the high performance in the experiments above.
However, CNN+Max does not capture relationships between piece images, which are important in tasks such as the shape composition task.
Even if CNN+Max is trained on more piece images, its performance does not improve.
On the other hand, CNN+MLP and CNN+GloRe learn how to combine piece images.
We hypothesized that when CNN+MLP and CNN+GloRe are given different number of piece images in training, the models can combine more piece images.
The results of the following experiments support our hypothesis: $|2,3 \rightarrow 4|$, $|1,2,3 \rightarrow 4|$ and $|1,2 \rightarrow 3,4|$.

\begin{figure*}[h]
    \begin{center}
    \includegraphics[scale=0.65]{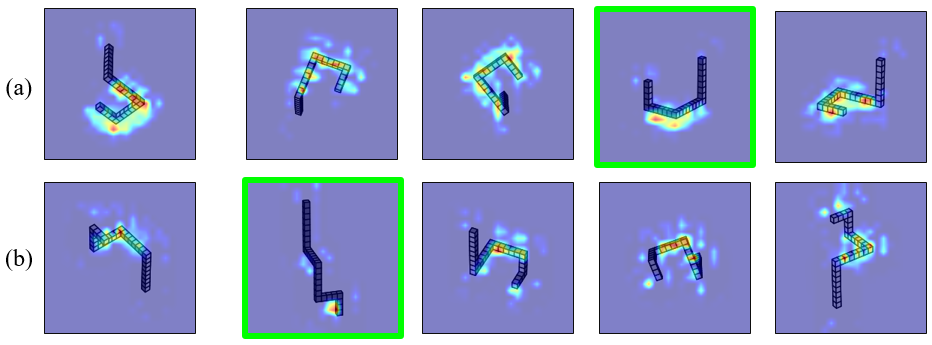}
    \caption{Grad-CAM visualization in the rotation task. The first column of each row denotes the question, and the last four images denote the candidates. Red and light regions denote part of objects that CNN+MLP captured. The correct answer is outlined with a green square. Best viewed in color.}
    \label{fig:correct_rotation_1}
    \end{center}
\end{figure*}

\section{Qualitative Analysis}
\label{section:qualitative}
In experiments, we confirmed that neural models can solve spatial reasoning tests, and generalize their ability even in the extrapolation setting.
However, it is still questionable whether they solve the tasks based on spatial understanding or pattern matching.
In this section, we provide further analysis and conclusions with visual aids.
For visualization, we utilize Grad-CAM \cite{selvaraju2017grad} which uses the gradient flows into the convolutional layer. We used the third convolutional layer of the CNNs to understand the important features for the answer.

\begin{figure*}[h]
    \begin{center}
    \includegraphics[scale=0.7]{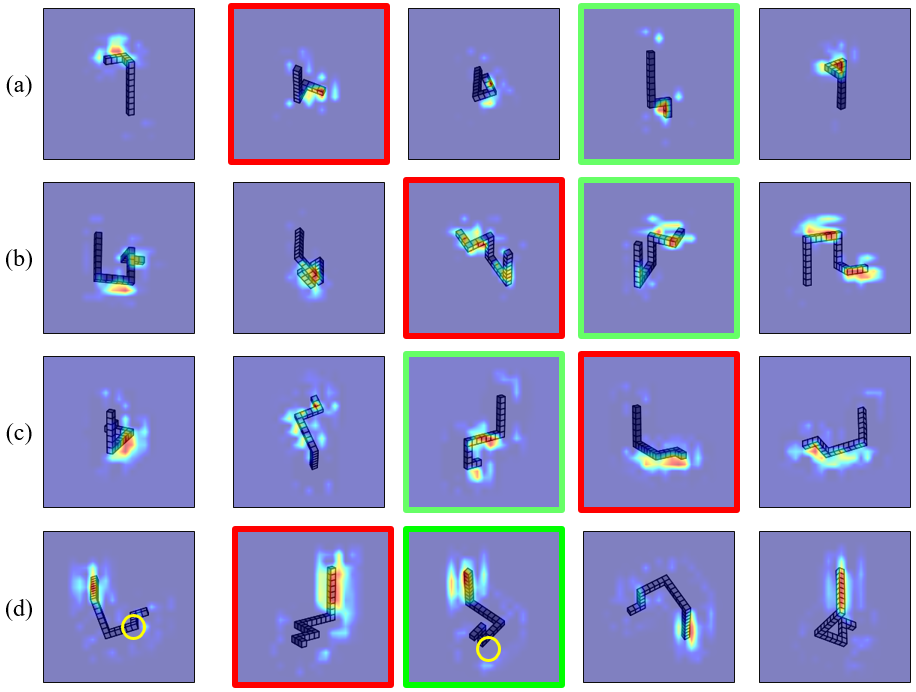}
    \caption{Error cases of CNN+MLP in the neutral setting. The first column of each row denotes the question, and the last four images denote the candidates. Red and light regions denote part of objects that CNN+MLP captured. The correct answer is outlined with a green square, and the incorrect model predictions are outlined with red squares. Best viewed in color. Analysis of the error cases is elaborated in Section \ref{section:qualitative}.}
    \label{fig:error_case_rotation}
    \end{center}
\end{figure*}

\subsection{Rotation}
We randomly sampled problems in the neutral setting in the rotation task that CNN+MLP predicted correctly, and analyzed them.
In most cases, we found that the model solves problems by capturing common structures of objects between question and candidates.
In Figure \ref{fig:correct_rotation_1}(a), the model captured the common '$\sqcup$'-shaped parts in the question and three candidates out of four and the L-shaped part in the remaining candidate.
As a result, the remaining candidate was chosen as the correct answer.
However, it is worth noting that the model did not always focus on '$\sqcup$'-shaped parts or L-shaped parts.
Its focus varies depending on objects.
The model occasionally captured joining blocks and the longest edges of objects.
Similarly, the model predicted the answer correctly when the model focused on the same parts in the question and the incorrect candidates, and different part in the correct candidate, as Figure \ref{fig:correct_rotation_1}(b) shows.

Next, we randomly sampled error cases of CNN+MLP, and classify our findings into three categories.
First, if the model captures the same structures from the question and all the candidates, i.e., there is no difference between candidates, it is confusing for the model.
In Figure \ref{fig:error_case_rotation}(a), the model focused on the same L-shaped parts from all the five objects, which results in incorrect prediction.
To solve the problem, the model is required to understand the direction to which the L-shaped parts are bent based on the longest edges.
The model failed to answer correctly due to the lack of spatial understanding, but humans can easily solve this problem.
Also, as Figure \ref{fig:error_case_rotation}(b) shows, the model is confused when the correct candidate answer image is similar to a mirror image of the other candidate.
Secondly, the model is vulnerable to the situation where a part of an object is obscured by another part of the object.
In Figure \ref{fig:error_case_rotation}(c), the L-shaped part in the question is obscured.
In this case, while human can restore the obscured part in their mind, the model recognizes only the frontmost part and misses the obscured part.
Lastly, despite the absence of obscurity, the model did not capture the important parts in solving problems.
In Figure \ref{fig:error_case_rotation}(d), we marked the important parts, the joining blocks, with yellow circles.
The model missed them, and mainly focused on the longest edges, which results in incorrect prediction.

\subsection{Shape Composition}

\begin{figure*}[h]
    \begin{center}
    \includegraphics[scale=0.65]{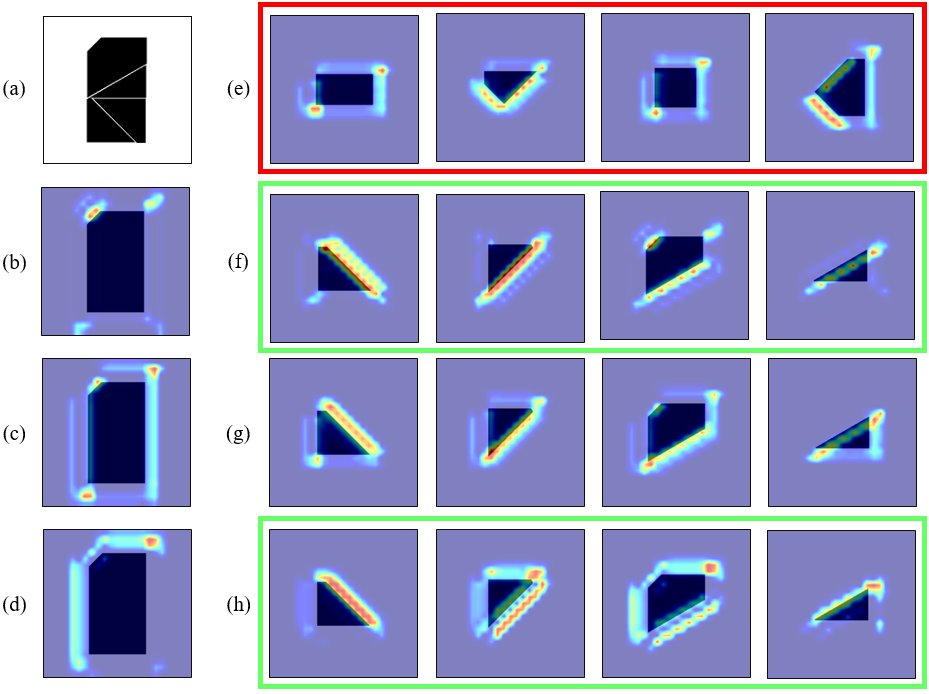}
    \caption{An example case where CNN+Max outperforms CNN+MLP$_{2}$. (a) Given question image with pieces drawn. (b),(c),(d): Grad-CAM visualization of the question image of CNN+Max, CNN+MLP$_{2}$ and CNN+MLP$_{1,2}$, respectively. (e) Incorrect answer for CNN+MLP$_{2}$. (f),(g),(h): Piece images from the correct answer for the corresponding model.}
    \label{fig:max_better}
    \end{center}
\end{figure*}

In this section, we analyze how neural models solve the shape composition task using Grad-CAM.
In the shape composition task, as Figure \ref{fig_data_example} shows, original images are adjacent to the border with no margins in the background. However, the weights are drawn outside the image outlines, thus information from the adjacent edges is lost resulting in difficulty of visual analysis.
Hence, we padded images with 50 margins at each side and re-scale the images to $224 \times 224$ size to prevent information loss.

Using newly preprocessed images, we trained CNN+MLP model and CNN+Max in $|2 \rightarrow 3,4|$ to investigate the reason of high performance of CNN+Max and the difference in problem solving between CNN+Max and CNN+MLP.\footnote{The performance of models after padding and re-scaling is slightly higher than before. In $|2 \rightarrow 3,4|$, CNN+MLP$_2$ and CNN+Max achieved 52.6\% and 65.1\%, respectively. CNN+MLP$_{1,2}$ achieved 68.3\%.}
We refer to the CNN+MLP model as CNN+MLP$_2$ to distinguish it from CNN+MLP$_{1,2}$ that is a CNN+MLP model trained on complexity level 1 and 2.
Note that CNN+Max first captures features of each piece image, and then selects the distinguishable ones among the features.
If the piece images have the same part of those features, the model predicts the candidate image as the answer.
For example, in Figure \ref{fig:max_better}(b) and \ref{fig:max_better}(f), CNN+Max captured an oblique side and a vertex on the top from the question image and captured the same part in the third piece image.
The CNN recognizer also captured other features, but max-pooling does not consider the combination of these features.
This problem solving method of CNN+Max is more effective than that of CNN+MLP$_2$ where the model rarely learns how to combine piece images, i.e., the model is trained on only one complexity level, or a single number of piece images.
In fact, CNN+MLP$_2$ did not solve the problem (Figure \ref{fig:max_better}(e) and (g)) because the model was trained to combine only two pieces, the model was not able to know how to generalize their ability to combine more than two pieces.

\begin{figure*}[h]
    \begin{center}
    \includegraphics[scale=0.65]{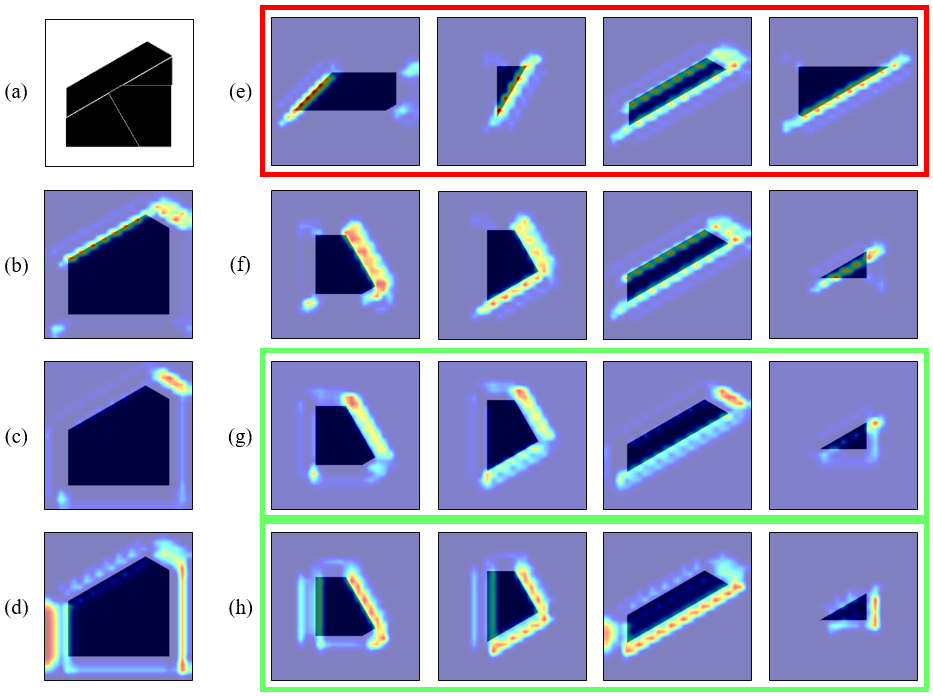}
    \caption{An example case where CNN+MLP$_{2}$ outperforms CNN+Max. (a) Given question image with pieces drawn. (b),(c),(d): Grad-CAM visualization of the question image of CNN+Max, CNN+MLP$_{2}$ and CNN+MLP$_{1,2}$, respectively. (e) Incorrect answer for CNN+Max. (f),(g),(h): Piece images from the correct answer for the corresponding model.}
    \label{fig:fc_better}
    \end{center}
\end{figure*}

However, CNN+Max is vulnerable to when piece images are rotated.
Since standard CNN filters do not consider rotation, CNNs capture different features from two images that are the same each other except that one of them is just rotated.
Thus, if the third piece image in Figure \ref{fig:max_better}(f) is rotated, the model outputs another answer.
CNN+Max is also confused when features from the question image are common across more than two candidate images.
In Figure \ref{fig:fc_better}, the top part of the question image (b) is very similar to the third piece image in the correct answer (f), but the model failed to answer correctly since there is the piece image in another candidate image (e) that is similar to the question image.
On the other hand, CNN+MLP$_{2}$ does not relies only on the similarity between the question and piece images and consider the combination of piece images.
As a result, CNN+MLP$_{2}$ solved the problem correctly, though similar piece images existed.

In addition, we trained another CNN+MLP model (CNN+MLP$_{1,2}$) in $|1,2 \rightarrow 3,4|$ to see the effect of training on different number of piece images.
Compared to CNN+MLP$_{2}$, CNN+MLP$_{1,2}$ captures the shapes of images more globally as Figure \ref{fig:max_better}(d), \ref{fig:max_better}(h), \ref{fig:fc_better}(d) and \ref{fig:fc_better}(h) shown.
This implies that CNN+MLP$_{1,2}$ consults more shape features prior to combining pieces.

There are several works studying the recognition ability of CNNs to capture image shapes \cite{kubilius2016deep,baker2018deep,geirhos2018imagenet}, but they did not provide visual analysis.
Unlike the finding that ImageNet-trained CNNs are insensitive to image shapes \cite{baker2018deep,geirhos2018imagenet}, our models solve the task based on information about image shapes.
However, as consistent with \cite{baker2018deep}, our CNN-based models usually do not capture global shapes of images, while it is natural for humans to capture global shapes when solving a Tangram puzzle.
Reducing this difference between humans and machines could be the key to solving puzzle-related tasks and even reasoning tasks.

\section{Conclusions and Future Work}
In this paper, we introduced two spatial reasoning test: rotation and shape composition, both of which are human IQ tests that require spatial reasoning.
We generated a dataset for each task, both of which consist of various complexity levels.
In experiments of the neutral and extrapolation settings, we examined whether neural net based models can apply their spatial reasoning ability to unseen situations, and confirmed that the models can do.
Several factors improve models' generalization: training on complex problems, training on different complexity levels and using reasoning modules such as MLPs.
Surprisingly, the max-pooling is effective in the extrapolation setting.
Another lesson is that higher performance in in-distribution does not guarantee better generalization.
Large model size and additional components may improve performance in in-distribution, but they can cause overfitting.
Also, we analyzed how baseline models solve spatial reasoning tests.
Although spatial reasoning tests were designed to measure spatial understanding, the models solve the tasks based on pattern matching with a lack of understanding of space.

Future work should focus on training models to understand the space.
If such training is possible, we can create a model that can solve complex problems just by learning simple problems.
Moreover, we can train our model efficiently with a small amount of data.

We simplified our tasks to focus on the reasoning abilities of neural models.
Based on the results from this study, we plan to extend these problem tasks to more general situations.
In the extended rotation task, polyominoes can be converted to cylindrical shapes.
In the extended shape composition task, the original shape and piece images can be converted to three dimensions.
In these extended tasks, more interesting and meaningful discoveries can be obtained.

{\small
\bibliographystyle{ieee_fullname}
\bibliography{egbib}

\begin{thebibliography}{10}\itemsep=-1pt

\bibitem{anderson2018vision}
Peter Anderson, Qi Wu, Damien Teney, Jake Bruce, Mark Johnson, Niko
  S{\"u}nderhauf, Ian Reid, Stephen Gould, and Anton van~den Hengel.
\newblock Vision-and-language navigation: Interpreting visually-grounded
  navigation instructions in real environments.
\newblock In {\em Proceedings of the IEEE Conference on Computer Vision and
  Pattern Recognition}, pages 3674--3683, 2018.

\bibitem{baker2018deep}
Nicholas Baker, Hongjing Lu, Gennady Erlikhman, and Philip~J Kellman.
\newblock Deep convolutional networks do not classify based on global object
  shape.
\newblock {\em PLoS computational biology}, 14(12):e1006613, 2018.

\bibitem{barrett2018measuring}
David~GT Barrett, Felix Hill, Adam Santoro, Ari~S Morcos, and Timothy
  Lillicrap.
\newblock Measuring abstract reasoning in neural networks.
\newblock {\em arXiv preprint arXiv:1807.04225}, 2018.

\bibitem{carlucci2019domain}
Fabio~M Carlucci, Antonio D'Innocente, Silvia Bucci, Barbara Caputo, and
  Tatiana Tommasi.
\newblock Domain generalization by solving jigsaw puzzles.
\newblock In {\em Proceedings of the IEEE Conference on Computer Vision and
  Pattern Recognition}, pages 2229--2238, 2019.

\bibitem{chen2019self}
Ting Chen, Xiaohua Zhai, Marvin Ritter, Mario Lucic, and Neil Houlsby.
\newblock Self-supervised gans via auxiliary rotation loss.
\newblock In {\em Proceedings of the IEEE Conference on Computer Vision and
  Pattern Recognition}, pages 12154--12163, 2019.

\bibitem{chen2019graph}
Yunpeng Chen, Marcus Rohrbach, Zhicheng Yan, Yan Shuicheng, Jiashi Feng, and
  Yannis Kalantidis.
\newblock Graph-based global reasoning networks.
\newblock In {\em Proceedings of the IEEE Conference on Computer Vision and
  Pattern Recognition}, pages 433--442, 2019.

\bibitem{cheng2016rifd}
Gong Cheng, Peicheng Zhou, and Junwei Han.
\newblock Rifd-cnn: Rotation-invariant and fisher discriminative convolutional
  neural networks for object detection.
\newblock In {\em Proceedings of the IEEE Conference on Computer Vision and
  Pattern Recognition}, pages 2884--2893, 2016.

\bibitem{clements2004geometric}
Douglas~H Clements.
\newblock Geometric and spatial thinking in early childhood education.
\newblock {\em Engaging young children in mathematics: Standards for early
  childhood mathematics education}, pages 267--297, 2004.

\bibitem{doersch2015unsupervised}
Carl Doersch, Abhinav Gupta, and Alexei~A Efros.
\newblock Unsupervised visual representation learning by context prediction.
\newblock In {\em Proceedings of the IEEE International Conference on Computer
  Vision}, pages 1422--1430, 2015.

\bibitem{ehrlich2006importance}
Stacy~B Ehrlich, Susan~C Levine, and Susan Goldin-Meadow.
\newblock The importance of gesture in children's spatial reasoning.
\newblock {\em Developmental psychology}, 42(6):1259, 2006.

\bibitem{gardner1992multiple}
Howard Gardner et~al.
\newblock {\em Multiple intelligences}, volume~5.
\newblock Minnesota Center for Arts Education, 1992.

\bibitem{geirhos2018imagenet}
Robert Geirhos, Patricia Rubisch, Claudio Michaelis, Matthias Bethge, Felix~A
  Wichmann, and Wieland Brendel.
\newblock Imagenet-trained cnns are biased towards texture; increasing shape
  bias improves accuracy and robustness.
\newblock {\em arXiv preprint arXiv:1811.12231}, 2018.

\bibitem{gidaris2018unsupervised}
Spyros Gidaris, Praveer Singh, and Nikos Komodakis.
\newblock Unsupervised representation learning by predicting image rotations.
\newblock {\em arXiv preprint arXiv:1803.07728}, 2018.

\bibitem{gur2017square}
Shir Gur and Ohad Ben-Shahar.
\newblock From square pieces to brick walls: The next challenge in solving
  jigsaw puzzles.
\newblock In {\em Proceedings of the IEEE International Conference on Computer
  Vision}, pages 4029--4037, 2017.

\bibitem{he2016deep}
Kaiming He, Xiangyu Zhang, Shaoqing Ren, and Jian Sun.
\newblock Deep residual learning for image recognition.
\newblock In {\em Proceedings of the IEEE conference on computer vision and
  pattern recognition}, pages 770--778, 2016.

\bibitem{hoshen2017iq}
Dokhyam Hoshen and Michael Werman.
\newblock Iq of neural networks.
\newblock {\em arXiv preprint arXiv:1710.01692}, 2017.

\bibitem{johnson2017clevr}
Justin Johnson, Bharath Hariharan, Laurens van~der Maaten, Li Fei-Fei, C
  Lawrence~Zitnick, and Ross Girshick.
\newblock Clevr: A diagnostic dataset for compositional language and elementary
  visual reasoning.
\newblock In {\em Proceedings of the IEEE Conference on Computer Vision and
  Pattern Recognition}, pages 2901--2910, 2017.

\bibitem{kanezaki2018rotationnet}
Asako Kanezaki, Yasuyuki Matsushita, and Yoshifumi Nishida.
\newblock Rotationnet: Joint object categorization and pose estimation using
  multiviews from unsupervised viewpoints.
\newblock In {\em Proceedings of the IEEE Conference on Computer Vision and
  Pattern Recognition}, pages 5010--5019, 2018.

\bibitem{kingma2014adam}
Diederik~P Kingma and Jimmy Ba.
\newblock Adam: A method for stochastic optimization.
\newblock {\em arXiv preprint arXiv:1412.6980}, 2014.

\bibitem{kipf2016semi}
Thomas~N Kipf and Max Welling.
\newblock Semi-supervised classification with graph convolutional networks.
\newblock {\em arXiv preprint arXiv:1609.02907}, 2016.

\bibitem{koch2015siamese}
Gregory Koch, Richard Zemel, and Ruslan Salakhutdinov.
\newblock Siamese neural networks for one-shot image recognition.
\newblock In {\em ICML deep learning workshop}, volume~2, 2015.

\bibitem{kubilius2016deep}
Jonas Kubilius, Stefania Bracci, and Hans P~Op de Beeck.
\newblock Deep neural networks as a computational model for human shape
  sensitivity.
\newblock {\em PLoS computational biology}, 12(4):e1004896, 2016.

\bibitem{marande2007mitochondrial}
William Marande and Gertraud Burger.
\newblock Mitochondrial dna as a genomic jigsaw puzzle.
\newblock {\em Science}, 318(5849):415--415, 2007.

\bibitem{noroozi2016unsupervised}
Mehdi Noroozi and Paolo Favaro.
\newblock Unsupervised learning of visual representations by solving jigsaw
  puzzles.
\newblock In {\em European Conference on Computer Vision}, pages 69--84.
  Springer, 2016.

\bibitem{paumard2018image}
Marie-Morgane Paumard, David Picard, and Hedi Tabia.
\newblock Image reassembly combining deep learning and shortest path problem.
\newblock In {\em Proceedings of the European Conference on Computer Vision
  (ECCV)}, pages 153--167, 2018.

\bibitem{pich2018logical}
Albert Pich and Zoe Falomir.
\newblock Logical composition of qualitative shapes applied to solve spatial
  reasoning tests.
\newblock {\em Cognitive Systems Research}, 52:82--102, 2018.

\bibitem{santa2017deeppermnet}
Rodrigo Santa~Cruz, Basura Fernando, Anoop Cherian, and Stephen Gould.
\newblock Deeppermnet: Visual permutation learning.
\newblock In {\em Proceedings of the IEEE Conference on Computer Vision and
  Pattern Recognition}, pages 3949--3957, 2017.

\bibitem{selvaraju2017grad}
Ramprasaath~R Selvaraju, Michael Cogswell, Abhishek Das, Ramakrishna Vedantam,
  Devi Parikh, and Dhruv Batra.
\newblock Grad-cam: Visual explanations from deep networks via gradient-based
  localization.
\newblock In {\em Proceedings of the IEEE International Conference on Computer
  Vision}, pages 618--626, 2017.

\bibitem{shen2017patch}
Xu Shen, Xinmei Tian, Shaoyan Sun, and Dacheng Tao.
\newblock Patch reordering: A novelway to achieve rotation and translation
  invariance in convolutional neural networks.
\newblock In {\em Thirty-First AAAI Conference on Artificial Intelligence},
  2017.

\bibitem{sholomon2014generalized}
Dror Sholomon, Omid~E David, and Nathan~S Netanyahu.
\newblock A generalized genetic algorithm-based solver for very large jigsaw
  puzzles of complex types.
\newblock In {\em Twenty-Eighth AAAI Conference on Artificial Intelligence},
  2014.

\bibitem{suchan2019out}
Jakob Suchan, Mehul Bhatt, and Srikrishna Varadarajan.
\newblock Out of sight but not out of mind: An answer set programming based
  online abduction framework for visual sensemaking in autonomous driving.
\newblock {\em arXiv preprint arXiv:1906.00107}, 2019.

\bibitem{wang2016solving}
Huazheng Wang, Fei Tian, Bin Gao, Chengjieren Zhu, Jiang Bian, and Tie-Yan Liu.
\newblock Solving verbal questions in iq test by knowledge-powered word
  embedding.
\newblock In {\em Proceedings of the 2016 Conference on Empirical Methods in
  Natural Language Processing}, pages 541--550, 2016.

\bibitem{weiler2018learning}
Maurice Weiler, Fred~A Hamprecht, and Martin Storath.
\newblock Learning steerable filters for rotation equivariant cnns.
\newblock In {\em Proceedings of the IEEE Conference on Computer Vision and
  Pattern Recognition}, pages 849--858, 2018.

\bibitem{zhang2019raven}
Chi Zhang, Feng Gao, Baoxiong Jia, Yixin Zhu, and Song-Chun Zhu.
\newblock Raven: A dataset for relational and analogical visual reasoning.
\newblock In {\em Proceedings of the IEEE Conference on Computer Vision and
  Pattern Recognition}, pages 5317--5327, 2019.

\end{thebibliography}
}

\appendix

\end{document}